\documentclass[journal]{IEEEtran}
\usepackage{amsmath,graphicx}

\usepackage[utf8]{inputenc} 
\usepackage[T1]{fontenc}    
\usepackage{hyperref}       
\usepackage{url}            
\usepackage{booktabs}       
\usepackage{amsfonts}       
\usepackage{nicefrac}       
\usepackage{microtype}      
\usepackage{csvsimple}
\usepackage{amsopn}
\usepackage{amsthm}
\usepackage{etex}
\usepackage{graphicx}
\usepackage{endnotes}
\usepackage{hyperref}
\usepackage{epsfig,psfrag}
\usepackage{pst-all}
\usepackage{amssymb,amsfonts,upref,cite,epsf,color,bm}
\usepackage{graphicx}
\usepackage{color}
\usepackage{amsmath}
\usepackage{graphicx}
\usepackage{calc}
\usepackage{booktabs}
\usepackage{tikz}
\usetikzlibrary{positioning,shapes,arrows}

\usepackage{pgfplots}
\newcommand\defeq{:=}

\usepackage{algorithm}
\usepackage{algpseudocode}
\floatname{algorithm}{Algorithm}
\algnewcommand\algorithmicinput{\textbf{Input:}}
\algnewcommand\INPUT{\item[\algorithmicinput]}
\algnewcommand\algorithmicoutput{\textbf{Output:}}
\algnewcommand\OUTPUT{\item[\algorithmicoutput]}

\DeclareMathOperator{\supp}{supp}

\DeclareMathOperator*{\argmin}{arg\;min}

\newcommand\vect[1]{\mathbf #1}

\newcommand{\vx}{\vect{x}}

\newcommand{\mC}{\mathbf{C}}

\newcommand{\samplesize}{m}

\newcommand{\sparsity}{s}

\newcommand{\featurelen}{n}

\newcommand{\expect}{{\rm E}}

\usepackage{setspace}
\usetikzlibrary{arrows,positioning} 
\tikzset{
	>=stealth',
	punkt/.style={
		rectangle,
		rounded corners,
		draw=black, very thick,
		text width=3cm,
		minimum height=2em,
		text centered},
	pil/.style={
		->,
		thick,
		shorten <=2pt,
		shorten >=2pt,}
}

\title{An Information-Theoretic Approach to Personalized Explainable Machine Learning}

\author{Alexander Jung and Pedro H. J. Nardelli 
	\thanks{AJ is with the Department of Computer Science, Aalto University, Finland. 
		PHJN is with Lappeenranta-Lahti University of Technology, Finland.  This work 
		is partly supported by Academy of Finland via: (a) ee-IoT project n.319009, (b) 
		FIREMAN consortium CHIST-ERA/n.326270, and (c) EnergyNet Research 
		Fellowship n.321265/n.328869.}
}

\begin{document}
	\maketitle
\begin{abstract}
Automated decision making is used routinely throughout our every-day life. 
Recommender systems decide which jobs, movies, or other user profiles might 
be interesting to us. Spell checkers help us to make good use of language. Fraud 
detection systems decide if a credit card transactions should be verified more closely. 
Many of these decision making systems use machine learning methods that fit complex 
models to massive datasets. The successful deployment of machine learning (ML) 
methods to many (critical) application domains crucially depends on its explainability. 
Indeed, humans have a strong desire to get explanations that resolve the uncertainty 
about experienced phenomena like the predictions and decisions obtained from ML methods. 
Explainable ML is challenging since explanations must be tailored (personalized) to individual 
users with varying backgrounds. Some users might have received university-level 
education in ML, while other users might have no formal training in linear algebra. 
Linear regression with few features might be perfectly interpretable for the first 
group but might be considered a black-box by the latter. We propose a simple 
probabilistic model for the predictions and user knowledge. This model allows to 
study explainable ML using information theory. Explaining is here considered as 
the task of reducing the ``surprise'' incurred by a prediction. We quantify the 
effect of an explanation by the conditional mutual information 
between the explanation and prediction, given the user background. 
\end{abstract}

\section{Introduction}
\label{sec_intro}
Machine learning (ML) methods compute predictions for quantities of interest 
based on a statistical analysis of large amounts of historical data \cite{hastie01statisticallearning,JungCompML,BishopBook}. 
These methods are routinely used to power many services within our everyday-life. 
ML methods power recommendation systems that decide what job ads or which other 
user profiles could be interesting to us \cite{Wang2018,Martinez2009}. 
Recent breakthroughs in ML, such as in image or text processing \cite{Goodfellow-et-al-2016}, 
also holds the promise of boosting the level of automation in domains which 
currently rely mainly on human labour or manual design \cite{Goodall2016}. 
 
%
 
A key challenge for the successful and ethically sound deployment of ML methods 
to critical application domains is the (lack of) explainability of its predictions 
\cite{Holzinger2018,Hagras2018,Mittelstadt2016,Wachter2017}. Explanations of 
predictions, which are used for decisions that affect humans, are increasingly 
becoming a legal obligation \cite{Wachter2017}. Beside legal aspects, it also seems that humans 
have a basic need for understanding decision making processes \cite{Kagan1972,Kruglanski1996}.  

One reason why explainable ML is challenging is that (good) explanations must be tailored 
to the knowledge of individual users (``explainee''). In general, for a particular prediction, 
there is no unique explanation that serves equally well a large group of heterogeneous users. 
Thus, achieving explainable ML would be easier for applications involving a homogenous group of 
users, like graduate students in a university program.

Large-scale applications as, for instance, recommendation systems for video streaming 
providers typically involve users with very different backgrounds, which can range from 
graduate studies in ML-related fields to users with no formal training in linear algebra. 
While linear models involving few hand-crafted features might be viewed as interpretable 
for the former group it might be considered a ``black-box'' for the latter group of users. 

This contribution studies explainable ML within information theory by using a probabilistic 
model for the data and user background. Loosely speaking, we model the 
effect of providing an explanation for a prediction as a reduction of 
the ``surprise'' incurred by a prediction to the user. This qualitative interpretation 
of explaining a prediction leads naturally to measuring the quantitative 
effect of explanations via (conditional) mutual information (MI) between the 
explanation and the prediction, given the user background (see Section \ref{sec_setup}). 

Our approach is different from existing work on explainable ML in the sense that 
we explicitly model the specific knowledge of each individual user. In contrast, 
most existing methods for explainable ML do not make any assumption about 
the end-user and her background knowledge.

Explainable ML methods can be roughly divided into two groups. The first group of 
methods uses models that are considered as intrinsically interpretable, like 
linear regression or small decision trees. The second group of methods, referred to as model-agnostic 
methods, probe an ML method by perturbing the features of the data point. 

The most straightforward approach to explainable ML methods is to use models 
that are considered to be intrinsically interpretable. Such methods include linear 
models, decision trees and artificial neural networks \cite{Montavon2018,Bach2015,Hagras2018}. 
Explaining the predictions obtained from such intrinsically interpretable models 
merely amounts to specifying the model parameters, such as the weights $w_{i}$ 
of a linear predictor $h(\mathbf{x}) = \sum_{i} w_{i} x_{i}$, or the feature-wise 
thresholds used in decision trees \cite{hastie01statisticallearning}. 

Interpretable models offer an intuitive decomposition of its predictions into a 
combination of elementary properties of a data point. Defining elementary 
properties of a data point via the activations of a (deep) neural network 
renders those models also interpretable (see \cite{Montavon2018}). 

Explainable models for sequential decision making have been studied in \cite{Mcinerney18}, 
where the authors obtain an explainable multi-armed bandit model by using 
the choice for the action space as the explanation. An explanation can be obtained 
by notifying the user that only previously purchased items are recommended. 
In contrast to \cite{Mcinerney18}, our approach uses a probabilistic model for 
the user background to compute personalized explanations that 
are optimal in a precise (information-theoretic) sense.  

A second group of explainable ML, referred to as model agnostic methods, is based on 
constructing explanations by probing a predictor as a black box \cite{Ribeiro2018,Hagras2018}. 
These methods aim at locally approximating black box models by simpler and interpretable 
models, such as linear models or shallow decision trees \cite{Ribeiro2018}. 

Our approach is also model agnostic as it only requires the statistical distributions 
of the model prediction. However, in contrast to most model agnostic explainable ML, 
we do not use local approximations to explain a black box method. Instead, 
we use a probabilistic model for the predictions and user knowledge. 

We frame explainable ML within a probabilistic model for ML predictions and 
user knowledge. This allows to capture the act of explaining a prediction using 
information-theoretic concepts. The act of explaining provides the user 
additional information about the prediction delivered by some (arbitrary) ML method. 

Information theory has already been used for learning 
optimal explanations \cite{Chen2018}. In a similar spirit, we also use 
MI to guide the learning of instance-wise explanations. However, 
in contrast to \cite{Chen2018}, we also model the effect of 
the user background on the information provided by an explanation. 
In a nutshell, while \cite{Chen2018} uses unconditional MI between explanations 
and predictions, we use the conditional MI given the user knowledge 
(see Section \ref{sec_optimal_explanation}). 
 
{\bf Outline and Contribution.}
In Section \ref{sec_setup}, we propose a simple probabilistic model for the features, 
prediction and user summary of a data point. This probabilistic model allows to quantify 
the effect of explanations via the conditional between the explanation 
and the model prediction, given the user background. 

Our main contribution is the formulation of an information-theoretic concept of optimal 
personalized explanations. As discussed in Section \ref{sec_optimal_explanation}), we construct  
(information-theoretically) optimal personalized explanations by maximizing the conditional 
MI between explanation and predictions, when conditioning on the user summary of a data point. 
To the best of our knowledge, we present the first information-theoretic approach to personalized 
explainable ML. 

A simple algorithm for computing optimal explanation given the 
model predictions and user summaries based on i.i.d. samples is presented in 
Section \ref{sec_simple_XML}. The proposed algorithm allows to construct 
personalized explanations that are optimal in an information-theoretic sense. 

\vspace*{-3mm}
\section{Problem Setup}
\label{sec_setup}
We consider a supervised ML problem involving data points with features 
$\vx = \big(x_{1},\ldots,x_{\featurelen}\big)^{T} \in \mathbb{R}^{\featurelen}$ 
and label $y \in \mathbb{R}$. Given some 
labelled training data 
\begin{equation}
\big(\mathbf{x}^{(1)},y^{(1)}\big),\big(\mathbf{x}^{(2)},y^{(2)}\big),\ldots,\big(\mathbf{x}^{(\samplesize)},y^{(\samplesize)}\big), 
\end{equation}
ML methods typically learn a predictor (map) 
\begin{equation} 
h(\cdot): \mathbb{R}^{\featurelen} \rightarrow \mathbb{R}: \vx \mapsto \hat{y}=h(\vx)
\end{equation}  
by requiring $\hat{y}^{(i)} \approx y^{(i)}$ \cite{hastie01statisticallearning,BishopBook,JungCompML}. 
\begin{figure}[htbp]
	\hspace*{5mm}
	\begin{tikzpicture}[node distance=1cm]
	\coordinate (OR) at (0.00, 1.50);
	\node[punkt] (data) {user $u$ consumig prediction $\hat{y}$};
	\node[above=of data] (dummy) {};
	\node[punkt,above=1cm of dummy] (hypothesis) {ML method};
	\node[right=1.4cm of dummy] (t) {prediction $\hat{y}$} ; 
	\node[left=1.4cm of dummy] (g) {explanation $e$} ; 
	\draw [->,line width=0.5mm] (hypothesis.east) to [out=0,in=90] (t.north);
	\draw [->,line width=0.5mm] (t.south) to [out=270,in=0] (data.east);
	\draw [<-,line width=0.5mm] (data.west) to [out=180,in=270] (g.south);
	\draw [<-,line width=0.5mm] (g.north) to [out=90,in=180] (hypothesis.west);
	\end{tikzpicture}
	\caption{An explanation $e$ provides additional information $I (\hat{y},e|u)$ to a user $u$ 
		about the prediction $\hat{y}$.}
	\label{fig_explainable_ML}
\end{figure}
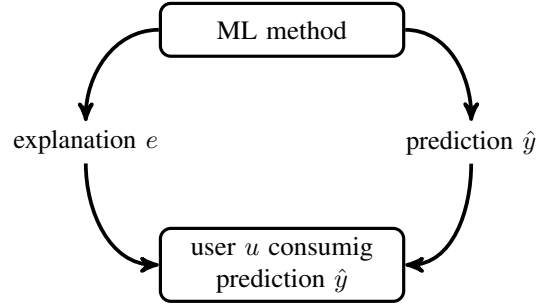

After learning a predictor $\hat{y}=h(\vx)$, it is applied to new data points 
yielding the prediction $\hat{y} = h(\vx)$. In may application, the prediction 
$\hat{y}$ is then delivered to a human user. The user can be the subscriber 
of a streaming service \cite{GomezUribe2016}, a dermatologist \cite{Esteva2017} 
or a city planner \cite{Yang2019}. 

Each user has typically some conception or model for the relation between features 
$\vx$ and label $y$ of a data point. Based on the user background, she has 
some understanding of a data point with features $\mathbf{x}$. 
 
Our approach to explainable ML is based on modelling the user understanding of 
a data point by some summary $u \in \mathbb{R}$. The summary is obtained by a 
(stochastic) map from the features $\vx$ of a data point. We will focus on summaries 
being obtained by a deterministic map
\begin{equation}
u(\cdot): \mathbb{R}^{\featurelen} \rightarrow \mathbb{R}: \mathbf{x} \mapsto u \defeq u(\mathbf{x}).
\end{equation} 
However, our approach also covers stochastic maps characterized by a conditional 
probability distribution $p(u| \vx)$. 

The (user-specific) quantity $u$ represents the understanding of the specific properties 
of the data point given the user knowledge (modelling assumptions). We interpret $u$ as 
a ``summary'' of the data point based on its features $\vx$ and the intrinsic modelling 
assumptions of the user. 

Let us illustrate the concept of the user summary $u$ as a means to represent 
user knowledge (or background) by two particular choices for $u$. First, the 
user summary could be the prediction obtained from a simplified model, such 
as linear regression using few features that the user anticipates as being relevant. 
Another example for a user summary $u$ could be a higher-level feature, such 
as eye spacing in facial pictures \cite{Jeong2015}. 

We formalize the act of explaining a prediction $\hat{y} = h(\vx)$ as presenting 
some additional quantity $e$ to the user. This ``explanation'' $e$ can be any 
quantity that helps the user to understand the prediction $\hat{y}$, given her 
understanding $u$ of the data point. Loosely speaking, the explanation $e$ 
contributes to resolving the uncertainty of the user $u$ about the prediction $\hat{y}$ \cite{Kagan1972}.  

For the sake of exposition, our focus will be on explanations obtained via 
a deterministic map 
\begin{equation}
\label{equ_def_explanation}
e(\cdot): \mathbb{R}^{\featurelen} \rightarrow \mathbb{R}: \vx \mapsto e \defeq e(\mathbf{x}), 
\end{equation} 
from the features $\vx$ of a data point. However, our approach can be generalized 
without difficulty to handle explanations obtained by a (stochastic) map. In the end, 
we only require the specification of the conditional probability distribution $p(e|\vx)$. 

Explanations can be constructed in quite different ways. An explanation could be a subset of 
features of a data point (see \cite{Ribeiro2016} and Section \ref{sec_optimal_explanation}). 
More generally, explanations could be obtained from simple local statistics (averages) of 
features that are considered related, such as near-by pixels in an image or consecutive samples 
of an audio signal. Instead of individual features, carefully chosen data points can also serve 
as an explanation \cite{Mcinerney18,Ribeiro2018}. 

To obtain comprehensible explanations that can be computed efficiently, 
we must typically restrict the space of possible explanations to a small subset $\mathcal{F}$ of maps 
\eqref{equ_def_explanation}. This is conceptually similar to the restriction of the space of possible 
predictor functions in a ML method to a small subset of maps which is known as the hypothesis space.

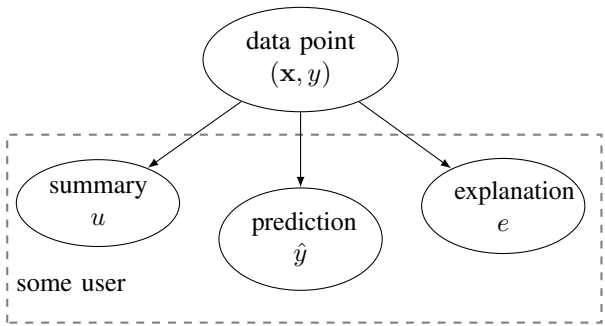
\begin{figure}[htbp]
	\hspace*{0mm}
\begin{tikzpicture}[
node distance=1cm and 0cm,
mynode/.style={draw,ellipse,text width=1.3cm,align=center}, 
mynode1/.style={draw,ellipse,text width=1.6cm,align=center}
]
\node[mynode1] (dp) {data point $(\mathbf{x},y)$};
\draw [color=gray,thick,dashed](-3.9,-3.5) rectangle (4,-1.0);
\node at (-3.9,-3) [above=1mm, right=0mm] {some user};
\node[mynode,below right=1cm and 1cm of dp] (exp) {explanation $e$};
\node[mynode,below left=1cm and 1cm of dp] (user) {summary $u$};
\node[mynode,below=of dp] (pred) {prediction $\hat{y}$};
\path (dp) edge[-latex] (exp)
(dp) edge[-latex] (user) ; 

\draw (dp) edge[-latex] (pred);
\end{tikzpicture}
	\vspace*{0mm}
	\caption{A simple probabilistic model for explainable ML.}
	\label{fig_simple_prob_ML}
\end{figure}

We consider data points as independent and identically distributed (i.i.d.) realizations of a random 
variable with fixed underlying probability distribution $p(\mathbf{x},y)$. Modelling the data point as 
random implies that the user summary $u$, prediction $\hat{y}$ and explanation $e$ are also random variables. 
The joint distribution $p(u,\hat{y},e,\mathbf{x},y)$ conforms with the Bayesian network \cite{Pearl1988} 
(depicted in Figure \ref{fig_simple_prob_ML}) since 
\begin{equation} 
\label{equ_joint_prob_factor}
p(u,\hat{y},e,\mathbf{x},y) = p(u|\mathbf{x}) \cdot p(e|\mathbf{x})  \cdot p(\hat{y}|\mathbf{x}) \cdot p(\vx,y). 
\end{equation}

We measure the amount of additional information provided by an explanation $e$ for 
a prediction $\hat{y}$ to some user $u$ via the conditional MI \cite[Ch. 2 and 8]{coverthomas}
\begin{equation} 
\label{eq_def_surprise}
I(e;\hat{y}|u) \defeq  \expect \bigg\{ \log \frac{p( \hat{y},e|u)}{p(\hat{y}|u)p(e|u)} \bigg\}.  
\end{equation} 
The conditional MI $I(e;\hat{y}|u)$ can also be interpreted as a measure for 
the amount by which the explanation $e$ reduces the uncertainty about the 
prediction $\hat{y}$ which is delivered to some user $u$. Thus, constructing 
explanations via solving \eqref{eq_def_surprise} conforms with the apparent human need 
to understand observed phenomena, such as the predictions from a ML 
method \cite{Kagan1972}. 
 
\section{Optimal Explanations}
\label{sec_optimal_explanation} 

Capturing the effect of an explanation using the probabilistic model \eqref{eq_def_surprise} 
offers a principled approach to computing an optimal explanation $e$. We require the optimal 
explanation $e^{*}$ to maximize the conditional MI \eqref{eq_def_surprise} 
between the explanation $e$ and the prediction $\hat{y}$ conditioned on the user summary $u$ of 
the data point. 

Formally, an optimal explanation $e^{*}$ solves
\begin{equation}
\label{equ_opt_explanation}
I(e^{*};\hat{y}|u) =  \sup_{e \in \mathcal{F}} I(e;\hat{y}|u). 
\end{equation} 
The choice for the subset $\mathcal{F}$ of valid explanations offers a trade-off 
between comprehensibility, informativeness and computational cost incurred 
by an explanation $e^{*}$ (solving \eqref{equ_opt_explanation}).

The maximization problem \eqref{equ_opt_explanation} for obtaining optimal explanations 
is similar to the approach in \cite{Chen2018}. However, while \cite{Chen2018} uses the 
unconditional MI between explanation and prediction, \eqref{equ_opt_explanation} 
involves the conditional MI given the user summary $u$.  
 
Let us illustrate the concept of optimal explanations \eqref{equ_opt_explanation} 
using a linear regression method. We model the features $\vx$ as a realization of 
a multivariate normal random vector with zero mean and covariance matrix $\mC_{x}$, 
\begin{equation} 
\label{equ_feature_vector_Gaussian}
\vx \sim \mathcal{N}(\mathbf{0},\mC_{x}).
\end{equation} 
The predictor and the user summary are linear functions of the features,  
\begin{equation} 
\label{equ_pred_summary}
\hat{y} \defeq \mathbf{w}^{T} \mathbf{x} \mbox{, and } u \defeq \mathbf{v}^{T} \mathbf{x}. 
\end{equation}

We construct explanations via subsets of individual features $x_{i}$ 
that are considered most relevant for a user to understand the prediction $\hat{y}$ 
(see \cite[Definition 2]{Montavon2018} and \cite{Molnar2019}). 
Thus, we consider explanations of the form 
\begin{equation} 
\label{equ_def_explanation}
e \defeq \{ x_{i} \}_{i \in \mathcal{E}} \mbox{ with some subset } \mathcal{E} \subseteq \{1,\ldots,\featurelen\}. 
\end{equation}

The complexity of an explanation $e$ is measured by the number $|\mathcal{E}|$ 
of features that contribute to it. We limit the number 
of features contributing to an explanation by a fixed (small) sparsity level, 
\begin{equation} 
| \mathcal{E} | \leq \sparsity (\ll \featurelen). 
\end{equation}

Modelling the feature vector $\vx$ as Gaussian \eqref{equ_feature_vector_Gaussian} implies 
that the prediction $\hat{y}$ and user summary $u$ obtained from \eqref{equ_pred_summary} is jointly 
Gaussian for a given $\mathcal{E}$ \eqref{equ_def_explanation}. 
Basic properties of multivariate normal distributions \cite[Ch. 8]{coverthomas}, 
allow to develop \eqref{equ_opt_explanation} as 
\begin{align}
\label{equ_sup_mi_Gauss}
  \max_{\substack{\mathcal{E}  \subseteq \{1,\ldots,\featurelen\} \\ |\mathcal{E}|  \leq s }} & I(e;\hat{y}|u)  \nonumber \\ 
     &=  h(\hat{y}|u) - h(\hat{y}|u,\mathcal{E}) \nonumber \\[3mm]
     &= (1/2) \log \mC_{\hat{y}|u} - (1/2) \log \det \mC_{\hat{y}|u,\mathcal{E}} \nonumber \\[3mm]
     & =  (1/2) \log \sigma^2_{\hat{y}|u} - (1/2) \log  \sigma^{2}_{\hat{y}|u,\mathcal{E}}. 
\end{align}
Here, $\sigma^2_{\hat{y}|u}$ denotes the conditional variance of the prediction $\hat{y}$, 
conditioned on the user summary $u$. Similarly, $\sigma^{2}_{\hat{y}|u,\mathcal{E}}$ denotes the conditional 
variance of $\hat{y}$, conditioned on the user summary $u$ and the subset $\{x_r\}_{r \in \mathcal{E}}$ of features. 
The last step in \eqref{equ_sup_mi_Gauss} follows from the fact that $\hat{y}$ is a scalar 
random variable. 

The first component of the last expression in \eqref{equ_sup_mi_Gauss} 
does not depend on the choice $\mathcal{E}$ for the explanation $e$ (see \eqref{equ_def_explanation}). 
Therefore, the optimal choice $\mathcal{E}$ solves
\begin{equation}
\label{equ_sup_m_sigma}
\sup_{|\mathcal{E}| \leq \sparsity}  - (1/2) \log  \sigma^{2}_{\hat{y}|u,\mathcal{E}}. 
\end{equation} 
The maximization \eqref{equ_sup_m_sigma} is equivalent to 
\begin{equation} 
\label{equ_min_variance}
\inf_{|\mathcal{E}| \leq \sparsity}  \sigma^{2}_{\hat{y}|u,\mathcal{E}}. 
\end{equation}

In order to solve \eqref{equ_min_variance}, we relate the 
conditional variance $\sigma^{2}_{\hat{y}|u,\mathcal{E}}$ to a particular 
decomposition
\begin{equation} 
\label{equ_def_linear_model}
\hat{y} = \alpha u + \sum_{i \in \mathcal{E}} \beta_{i} x_{i} + \varepsilon. 
\end{equation}
For an optimal choice of the coefficients $\alpha$ and $\beta_{i}$, 
the variance of the error term in \eqref{equ_def_linear_model} is given by $\sigma^{2}_{\hat{y}|u,\mathcal{E}}$. 
Indeed, 
\begin{equation} 
\label{equ_def_optimal_coef_linmodel}
 \min_{\alpha,\beta_{i} \in \mathbb{R}} \expect\big\{ \big(\hat{y} - \alpha u - \sum_{i \in \mathcal{E}} \beta_{i} x_{i} \big)^{2}\big\} =  \sigma^{2}_{\hat{y}|u,e}. 
\end{equation} 

Inserting \eqref{equ_def_optimal_coef_linmodel} into \eqref{equ_min_variance}, an optimal 
choice $\mathcal{E}$ (of feature) for the explanation of prediction $\hat{y}$ to user $u$ 
is obtained from  
\begin{align} 
 & \inf_{|\mathcal{E}| \leq \sparsity}  \min_{\alpha,\beta_{i} \in \mathbb{R}} \expect\big\{ \big(\hat{y} - \alpha u - \sum_{i \in \mathcal{E}} \beta_{i} x_{i} \big)^{2}\big\}  \label{equ_final_opt_E} \\
  &= \min_{ \| {\bm \beta} \|_{0} \leq s }  \expect\big\{ \big(\hat{y} - \alpha u - {\bm \beta}^{T} \vx \big)^{2}\big\}   \label{equ_final_opt_beta}.
\end{align} 
An optimal subset $\mathcal{E}_{\rm opt}$ of features defining the explanation $e$ \eqref{equ_def_explanation}
is obtained from any solution ${\bm \beta}_{\rm opt}$ of \eqref{equ_final_opt_beta} via
\begin{equation}
\label{equ_opt_expl_support}
\mathcal{E}_{\rm opt} = \supp {\bm \beta}_{\rm opt}. 
\end{equation}

\section{A Simple XML Algorithm}
\label{sec_simple_XML}

Under a Gaussian model \eqref{equ_feature_vector_Gaussian} for the features of data points, 
Section \ref{sec_optimal_explanation} shows how to construct optimal explanations via the 
(support of the) solutions ${\bm \beta}_{\rm opt}$ of the sparse linear regression problem \eqref{equ_final_opt_beta}. 

In order to obtain a practical algorithm for computing (approximately) optimal explanations \eqref{equ_opt_expl_support}, 
we need to approximate the expectation in \eqref{equ_final_opt_beta} with an empirical 
average over i.i.d.\ samples $\big(\vx^{(i)},\hat{y}^{(i)},u^{(i)}\big)$ of features, predictions 
and user summaries. This results in Algorithm \ref{alg:xml}.

\begin{algorithm}[htbp]
	\caption{XML Algorithm}\label{alg:xml}
	\begin{algorithmic}[1]
		\renewcommand{\algorithmicrequire}{\textbf{Input:}}
		\renewcommand{\algorithmicensure}{\textbf{Output:}}
		\Require  explanation sparsity $\sparsity$, training samples $\big(\vx^{(i)},\hat{y}^{(i)},u^{(i)}\big)$ for $i=1,\ldots,\samplesize$
		\State compute $\widehat{\bm \beta}$ by solving 
	\begin{equation}
	\label{equ_P0}
\widehat{\bm \beta} \in  	 \argmin_{ \| {\bm \beta} \|_{0} \leq s }  \sum_{i=1}^{\samplesize}  \big(\hat{y}^{(i)} - \alpha u^{(i)} - {\bm \beta}^{T} \vx^{(i)} \big)^{2}
	\end{equation}
		\Ensure feature set $\widehat{\mathcal{E}} \defeq {\rm supp} \widehat{\bm \beta}$
	\end{algorithmic}
\end{algorithm}
Note that Algorithm \ref{alg:xml} is interactive since the user has to provide 
samples $u^{(i)}$ of its summary for the data points with features $\vx^{(i)}$. 
Based on the user input $u^{(i)}$, for $i=1,\ldots,\samplesize$, Algorithm \ref{alg:xml} 
learns an optimal subset $\mathcal{E}$ of features  \eqref{equ_def_explanation} that  
are used for the explanation of predictions. 

The sparse regression problem \eqref{equ_P0} becomes 
intractable for large feature length $\featurelen$. However, if the features are 
weakly correlated with each other and the user summary $u$, the solutions of 
\eqref{equ_P0} can be found by convex optimization. Indeed, for a wide range 
of settings, sparse regression \eqref{equ_P0} can be solved via a convex 
relaxation, known as the least absolute shrinkage and selection operator (Lasso) \cite{HastieWainwrightBook}, 
\begin{equation} 
\label{equ_Lasso}
\widehat{\bm \beta} \!\in\!\argmin_{ {\bm \beta} \in \mathbb{R}^{\featurelen}}  \sum_{i=1}^{\samplesize}  \big(\hat{y}^{(i)} - \alpha u^{(i)} - {\bm \beta}^{T} \vx^{(i)} \big)^{2} + \lambda \| {\bm \beta} \|_{1}.  
\end{equation}
We have already a good understanding of choosing the Lasso parameter 
$\lambda$ in \eqref{equ_Lasso} such that its solutions coincide with the 
solutions of \eqref{equ_P0} (see, e.g., \cite{HastieWainwrightBook}). 

\section{Numerical Experiments} 
\begin{figure}[h]
	\includegraphics[width=\columnwidth]{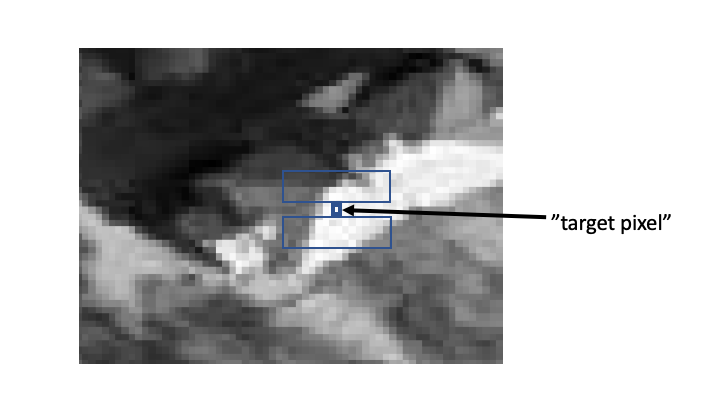}
	\vspace*{-11mm}
	\caption{The greyscale level of a particular target pixel within an aerial photograph can be predicted based on the greyscale values of nearby pixels within adjacent rectangles (indicated).}
	\label{fig:patch_target}
	\vspace*{-1mm}
\end{figure}

We verify the ability of Algorithm \ref{alg:xml} to provide explainable ML using 
a computer vision application. In particular, we consider data points representing 
square patches of a greyscale aerial photograph of Helsinki city area.\footnote{The 
	data is freely available via the online map service \url{https://kartta.hel.fi/}} The goal 
is to predict the greyscale value $y$ of the center (``target'') pixel. In order to predict 
the greyscale value of the $i$th pixel $y^{(i)}$ we use the greyscale values $x_{j}^{(i)}$ of 
close-by pixels $j \in \mathcal{P}^{(i)}$. As depicted in Figure \ref{fig:patch_target}, the 
neighbourhood $j \in \mathcal{P}^{(i)}$ is constituted by two rectangular areas that 
are adjacent to pixel $i$. 
A user having some prior experience in processing natural images might consider 
the average greyscale value 
\begin{equation}
u^{(i)} = (1/ | \mathcal{P}^{(i)}|)\sum_{j \in  \mathcal{P}^{(i)}} x_{j}^{(i)}
\end{equation}
as a reasonable summary of the features $\vx^{(i)} \defeq \{ x_{j}^{(i)}\}_{j \in \mathcal{P}^{(i)}}$. 
We refer to the Python notebook 
\url{https://github.com/alexjungaalto/ResearchPublic/blob/master/itxml.ipynb} 
for the results of the experiments. 

\section{Conclusion} 
We have introduced a simple probabilistic model for the predictions of 
a ML method and the user background. The user background is represented 
by a summary of the features of a data point. The effect of an explanation 
is measured by the conditional MI between prediction 
and explanation, given the user summary of a data point. 

\bibliographystyle{plain}
\bibliography{/Users/alexanderjung/Literature}

\end{document}